\pdfoutput=1

\documentclass[11pt]{article}

\usepackage[final]{acl}

\usepackage{times}
\usepackage{latexsym}
\usepackage{amsfonts}
\usepackage{booktabs,makecell, multirow, tabularx, enumitem}
\usepackage{adjustbox}
\usepackage{makecell}
\usepackage{enumitem}

\usepackage[T1]{fontenc}

\usepackage[utf8]{inputenc}

\usepackage{microtype}

\usepackage{inconsolata}

\usepackage{graphicx}
\usepackage{amsmath}
\usepackage{enumitem}
\usepackage{booktabs}
\title{\raisebox{-0.05cm}{\includegraphics[height=1\baselineskip]{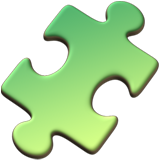}}\textsc{Personalized Pieces}: Efficient Personalized Large Language Models through Collaborative Efforts}

\author{Zhaoxuan Tan, Zheyuan Liu, Meng Jiang \\
        University of Notre Dame\\
        \texttt{\{ztan3, mjiang2\}@nd.edu}}

\newcommand{\ourmethod}{\textsc{Per-Pcs}}

\begin{document}
\maketitle

\begin{abstract}

Personalized large language models (LLMs) aim to tailor interactions, content, and recommendations to individual user preferences. While parameter-efficient fine-tuning (PEFT) methods excel in performance and generalization, they are costly and limit communal benefits when used individually. To this end, we introduce \textsc{Personalized Pieces} (\ourmethod{})\footnote{Code is available at \url{https://github.com/TamSiuhin/Per-Pcs}}, a framework that allows users to safely share and assemble personalized PEFT efficiently with collaborative efforts.
\ourmethod{} involves selecting sharers, breaking their PEFT into pieces, and training gates for each piece. These pieces are added to a pool, from which target users can select and assemble personalized PEFT using their history data. This approach preserves privacy and enables fine-grained user modeling without excessive storage and computation demands.
Experimental results show \ourmethod{} outperforms non-personalized and PEFT retrieval baselines, offering performance comparable to OPPU with significantly lower resource use across six tasks. Further analysis highlights \ourmethod{}'s robustness concerning sharer count and selection strategy, pieces sharing ratio, and scalability in computation time and storage space.
\ourmethod{}'s modularity promotes safe sharing, making LLM personalization more efficient, effective, and widely accessible through collaborative efforts.

\end{abstract}

\section{Introduction}
Personalization involves mining user's history data to tailor and customize a system's interaction, content, or recommendations to meet the specific needs, preferences, and characteristics, of individual users \cite{tan2023user, chen2023large, kirk2024benefits}. By adapting to each user's unique preferences, personalization enhances the user experience and has become increasingly important in content recommendation \cite{li2023text, wu2023personalized, baek2023knowledge}, user simulation \cite{dejescu2023approaches, zhang2020evaluating}, personalized chatbot \cite{srivastava2020personalized, ma2021one}, user profiling \cite{gu2020hierarchical, gao2023chat}, healthcare \cite{johnson2021precision, goldenberg2021personalization}, and education \cite{alamri2021learning, pratama2023revolutionizing}. 


\begin{figure}[t]
    \centering
    \includegraphics[width=1\linewidth]{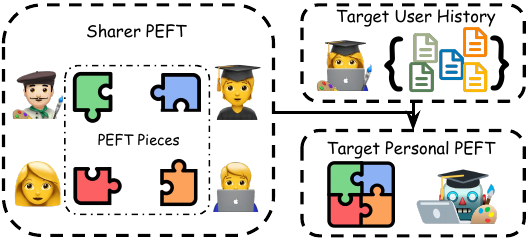}
    \caption{User personal PEFT parameter sharing framework: Sharers provide parts of their PEFT parameters (PEFT pieces). Using the target user's history data, we recycle the PEFT pieces shared by anchor users and assemble the target user's personal PEFT.}
    \label{fig:teaser}
\end{figure}

Large language models (LLMs) are revolutionizing the research landscape with emergent abilities not observed in smaller models \cite{wei2022emergent, lu2023emergent}, due to their training on massive textual corpora and billions of parameters. These abilities include step-by-step reasoning \cite{wei2022chain}, in-context learning \cite{min2022rethinking}, and instruction following \cite{wei2021finetuned}. Despite these capabilities, current LLMs adhere to a "one-size-fits-all" paradigm, being trained on broad, domain-agnostic data, which limits their effectiveness in adapting to individual user preferences \cite{chen2023large}. Consequently, personalizing LLMs to align with users' unique needs has become a crucial research focus \cite{li2023teach}.





Previous endeavors to personalize LLMs can be categorized into prompt-based and parameter-efficient fine-tuning (PEFT)-based methods. 
Prompt-based personalization involves designing prompt templates to help LLMs understand user preferences, using methods such as vanilla personalized prompting \cite{dai2023uncovering}, retrieval-augmented prompting \cite{mysore2023pearl}, and profile-augmented prompting \cite{richardson2023integrating}. However, prompt-based methods expose user data to centralized LLM and can be easily distracted by irrelevant user history data, which retrieval can hardly avoid \cite{shi2023large}. \textit{PEFT-based} personalization methods focus on storing users' preferences and behavior patterns in personal lightweight parameters. OPPU \cite{tan2024democratizing} is the pioneering work that stores users' preferences and behavior patterns in personal PEFT parameters, showing the superiority of model ownership and better user behavior pattern generalization compared to prompt-based methods.
Despite their success, the ``one-PEFT-per-user" paradigm is computationally and storage-intensive, especially for large user bases. For instance, using OPPU for personalized product rating prediction requires about 20 minutes of training on a single RTX A6000 GPU and 17 MB of storage per user, scaling linearly with the number of users. Additionally, individually owned PEFTs limit community value, as personal models cannot easily share knowledge or benefit from collaborative improvements.
Inspired by the exhaustiveness of human preferences \cite{lee2024aligning}, we propose the \textsc{Personalized Pieces} (\ourmethod{}) framework, which allows users to safely share a small fraction of their PEFT parameters and build personalized LLMs efficiently through collaborative efforts (Figure \ref{fig:teaser}).
Specifically, we first select representative users as sharers and train their PEFTs with their personal history data. We then break down the PEFT parameters into pieces, inject a routing gate for each piece, and update the gate parameters while keeping the other parameters frozen with a few steps. These pieces are added to a pieces pool along with their corresponding gates for selection.
In the assembly stage, \ourmethod{} feeds the target user's history data and selects PEFT pieces from the pieces pool in an auto-regressive way, recycling the PEFT modules in the pieces pool. By processing all the history data through this pipeline, we determine the PEFT piece choices for all layers and obtain the target user's personal PEFT. \ourmethod{} is training-free and only requires the storage of sharer's index and corresponding composition weights, making it computation and storage efficient. 
Experimental results show that \ourmethod{} outperforms non-personalized and PEFT retrieval baselines, delivering performance comparable to OPPU but with significantly reduced resource requirements across six personalization tasks in the LaMP benchmark \cite{salemi2023lamp}. Further studies highlight \ourmethod{}'s robustness against sharer count and selection strategy. Even when sharers consent to share only a small portion of their pieces, \ourmethod{} maintains strong performance, comparable to scenarios where all pieces are shared. Time analysis reveals that \ourmethod{} is 38 times more efficient in storage and 7 times more efficient in computation costs compared to OPPU. These findings underscore the potential of personalizing general-purpose LLMs by integrating modular and collaborative parametric knowledge from personal PEFT pieces shared by users.


In summary, the contribution of \ourmethod{} is the pioneering framework that enables users to safely share personal PEFTs, facilitating efficient and fine-grained LLM personalization through collaborative efforts. Unlike OPPU, where personal PEFTs benefit only the individual user, \ourmethod{} allows users to share a limited portion of their PEFT parameters with others, ensuring user privacy. For target users, \ourmethod{} maintains model ownership and supports fine-grained user modeling comparable to OPPU, but with significantly reduced storage and computation resources. We envision \ourmethod{} as an initiative to encourage users to share their personal PEFT pieces, fostering collaboration in personalizing LLMs to create value for others. This approach preserves sharer privacy and reduces the carbon footprint of PEFT-based personalized LLM.

\begin{figure*}[t]
    \centering
    \includegraphics[width=0.9\linewidth]{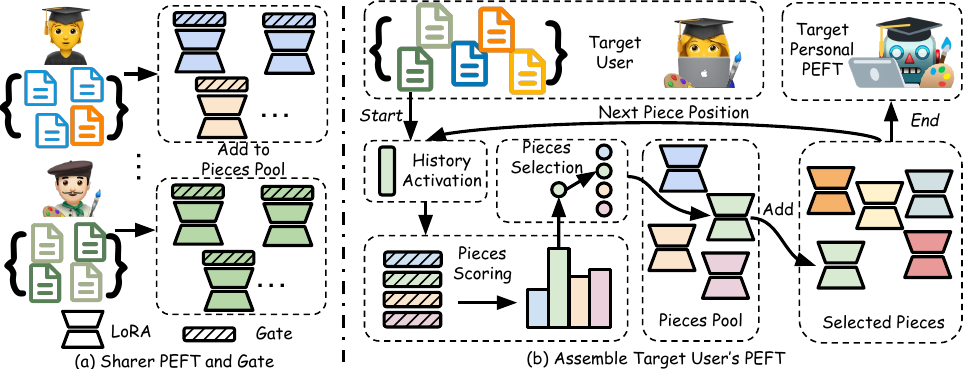}
    \caption{Overview of \ourmethod{}. First, we train PEFT and gate each piece for sharing. Next, we feed the target user's history, utilizing history activation and piece gates to score and select PEFT pieces from the pool. These selected pieces are then assembled to create a personalized PEFT for the target user.}
    \label{fig:overview}
\end{figure*}

\section{\textsc{Personalized Pieces} (\ourmethod{})}

We introduce \ourmethod{}, a novel framework to empower LLM personalization with modular and collaborative PEFT pieces within the community (Fig. \ref{fig:overview}). We first adapt non-personalized base LLMs to the task without incorporating personal preferences (\S \ref{sec:basellm}). We then train personal PEFT and post-hoc gates for sharers and add them to the pool (\S \ref{sec:anchorgate}). Finally, we assemble the target user's PEFT using their history and pieces from the pool (\S \ref{sec:assemble}).

\subsection{Preliminaries}
\textbf{Research Problem Formulation.} 
For personalized LLM at time $t$, the model's output $r_u$ for user $u$ is conditioned on both query $q_u$ and the user’s behavior history $\mathcal{H}_u = \{h_u\}$ that  includes all user behaviors occurred before query time $t$. Assuming users in set $\mathcal{U}$ have personal PEFT, while the target user $\hat{u}\notin \mathcal{U}$ does not, our goal is to assemble the target user's PEFT $\Delta\Theta_{\hat{u}}$ from $\{\Delta\Theta_{u}, u\in \mathcal{U}\}$.

\noindent\textbf{PEFT Pieces.} We assume a PEFT method introduces modules throughout the whole model. For example, LoRA \cite{hu2021lora} introduces a low-rank update at every linear layer in the model. We refer to each of these updates as a ``piece”.

\subsection{Base LLM Task Adaption}
\label{sec:basellm}
Since off-the-shelf LLMs do not inherently understand personalization tasks, we follow LaMP \cite{salemi2023lamp} and \citet{richardson2023integrating} to fine-tune LLMs for fair comparison and task comprehension. In adapting the base LLM, we use data that excludes target users' and sharers' data to build an LLM that understands task-related capabilities rather than personal preferences. Specifically, the base LLM parameter $\Theta_{o}$ is optimized \emph{w.r.t.} loss $L=\mathrm{CE}[\Theta_o(\phi(q_u, \mathcal{R}(q_u, \mathcal{H}_u, m))), r_u]$, where $\mathrm{CE}$ denotes the cross entropy loss function, $\mathcal{R}$ is the retriever, $\phi$ is the prompt construction function, $m$ is the number of retrieval items, and $\mathcal{H}_u$ is the entire user behavior history. For computational efficiency, we adopted LoRA \cite{hu2021lora} for parameter-efficient fine-tuning and merged it into pretrained weights to obtain the base LLM.

\subsection{Sharer Selection and PEFT Training}
\label{sec:anchorgate}
After adapting the base LLMs to the task without incorporating the target user's personal preferences, we select sharers who consent to share their PEFTs with the community and train their personal PEFTs. To select representative users, we first get embeddings for all candidate users by encoding their history with an encoder-only language model, \texttt{DeBERTa-v3-Large} \cite{he2022debertav3}. The user embedding $E_u=\sum_{h_u\in \mathcal{H}_u}\mathrm{Enc}(h_u)/{|\mathcal{H}_u|}$ by averaging all history items $h_u$ from user $u$. We then cluster user embeddings with the k-means algorithm ($K$=50 by default) and select the most active users within the $i$-th cluster as sharer $s_i$ $(i=1,..., K)$,\footnote{Please see more sharer selection strategies in Section \ref{sec:anchor_selection}.}
\begin{align*}
    s_i=\{\arg\max_{u\in \mathcal{C}_i}|\mathcal{H}_u|, \mathcal{C}_i\in\textit{k-means}(\mathcal{E}, K)\},
\end{align*}
where $\mathcal{C}_i$ denotes the $i$-th user cluster, $\mathcal{E}=\{E_u, u\in \mathcal{U}\}$ denotes the embedding set of all sharer candidates.
Following OPPU \cite{tan2024democratizing}, we then train personal PEFT parameters $\Theta_{s_i}$ for sharer $s_i$ using sharer's history data $\mathcal{H}_{s_i}$. 

We then break the sharers' PEFT parameters $\Theta_{s_i}$ into pieces. For clarity, we consider the case where users perform personal PEFT using LoRA \cite{hu2021lora}. It's worth noting that \ourmethod{} is compatible with all PEFT methods that introduce trainable modules throughout the model, such as Adapter \cite{houlsby2019parameter}, $\mathrm{(IA)^3}$ \cite{liu2022few}, and prefix tuning \cite{li-liang-2021-prefix}. We primarily focus on LoRA due to its popularity, widespread use, and superior performance demonstrated by OPPU. LoRA modifies the output of $l$-th linear layer from $z_t^l = W_o^l v_t^l$ using a low-rank decomposition to $z_t^l = W_o^l v_t^l + \Delta W^l v_t^l = W_o^l v_t^l + B^lA^l v_t^l$, where $v_t^l \in \mathbb{R}^{n}$ denotes the $t$-th input activation at layer $l$, $W_o^l \in \mathbb{R}^{d \times n}$ denotes the base model parameters which remain frozen during fine-tuning, $A^l \in \mathbb{R}^{r \times n}$ and $B^l \in \mathbb{R}^{d \times r}$ are trainable parameters. Therefore, a pair of $(B^l, A^l)$ is defined as a "piece" and personal PEFT parameters $\Theta_{s_i}$ for sharer $s_i$ can break down to pieces set $\{(B_{s_i}^l, A_{s_i}^l)\}_{l=1}^{L}$, $L$ denotes the total number of layers in a LoRA module.

\subsection{Post-Hoc Sharer Gating Training}

We then add a piece selection gate for each sharer PEFT piece to determine which piece should be selected in the upcoming assembly step. For each sharer PEFT piece, after integrating the gate, a linear layer becomes:
\begin{align*}
    z_t^l = W_o^l v_t^l + B_{s_i}^lA_{s_i}^l v_t^l \sigma (g_{s_i}^{l\top} v_t^l),
\end{align*}
where $g_{s_i}^l \in \mathbb{R}^n$ is a trainable gate vector for sharer $s_i$ at layer $l$ and initialized to all zeros, $\sigma$ is the sigmoid activation function, $W_o^l$, $B_{s_i}^l$, and $A_{s_i}^l$ are frozen. For sharer $s_i$, we optimize $\{g_{s_i}^l\}_{l=1}^L$ using sharer history $\mathcal{H}_{s_i}$. We then add the sharers' PEFT pieces and corresponding gates to the pieces pool for the upcoming selection and assembly. Gate training is limited to around 50 steps, making it computationally efficient. The post-hoc nature of gate learning also adds flexibility, facilitating easier deployment in real-world scenarios.

\subsection{Assemble Target Personal PEFT}
\label{sec:assemble}
Motivated by the exhaustiveness of human preferences \cite{lee2024aligning}, we assemble PEFT modules for target users using the PEFT modules and gate vectors from sharers. Using the target user's history $\mathcal{H}_{\hat{u}}$ as input, we perform auto-regressive PEFT piece selection to assemble the target user's PEFT from input to output. For each layer $l$ in LoRA, we feed the user history in LLM and compute the score for the input activation $v_t^l$ and candidate pieces using cosine similarity, then aggregate these scores from the token level to obtain the piece-level score $\alpha_{s_i}^l$ for the piece from sharer $s_i$:
\begin{align*}
\alpha_{s_i}^l = \sum_{t=b}^{e}({\overline{g}_{s_i}^{l\top} \overline{v}_t^l}), 
\end{align*}
where $\overline{g}_{s_i}^l$ and $\overline{v}_t^l$ are the normalized gate vector and activation. For user history $(x_{\hat{u}}, y_{\hat{u}}) \in \mathcal{H}_{\hat{u}}$ aligned with the task format, we set begin position $b=|x_{\hat{u}}|+1$ and end position $e=|x_{\hat{u}}|+|y_{\hat{u}}|+1$, where $|\cdot|$ denotes sequence length. Otherwise, for history $x_{\hat{u}} \in \mathcal{H}_{\hat{u}}$, we set $b=1$ and $e=|x_{\hat{u}}|+1$.

We then select the top-$k$ PEFT pieces at $l$-th layer to select sharer set $\mathcal{S}^l$ for target user PEFT assemble $\mathcal{S}^l=\{s_i \text{, keep top-$k$ ranked by }\alpha_{s_i}^l\}$.
Next, we normalize the selected weights with the $1/\sqrt{n}$ scaling factor to avoid saturation \cite{vaswani2017attention}, which can be expressed as
\begin{align*}
    w^l_{s} = \textit{softmax}(\{\alpha_{s}^l/ \sqrt{n}, \; s\in \mathcal{S}^l\}),
\end{align*}
where $n$ denotes the embedding dimension, $s$ denotes the index of selected pieces. We then aggregate the selected PEFT pieces with weight to assemble the target user's PEFT $\Delta W_{\hat{u}}^l$ at layer $l$:
\begin{align*}
    \Delta W_{\hat{u}}^l = \sum_{s\in \mathcal{S}^l}(w_{s}^l B_{s}^l A_{s}^l)
\end{align*}
where $A_{s}^l$ and $B_{s}^l$ are PEFT piece parameters from $s$-th sharer. Using the assembled personal PEFT parameters for target user $\hat{u}$, the feed forward function for a linear layer becomes
\begin{align*}
    z_t^l = W_o^l v_t^l + \Delta {W^l_{\hat{u}}} v_t^l.
\end{align*}

After detailing the assembly process for a single piece in the target user's PEFT, we extend it to the entire model that contains $L$ layers.
Once the piece at $l$-th layer parameter assembly is complete, the output $z_t^l$ is used as the input activation for the $l$+1 layer selection. After computing parameter selection for all history items and layers, we average the composed parameters to obtain the final PEFT parameters for the target user $\Delta\Theta_{\hat{u}}=\{\Delta W_{\hat{u}}^l\}_{l=1}^{L}$, which is a set of assembled parameters across all layers sourced from sharers' piece parameters.

Overall, the assembly process does not involve model training or optimization, making it computationally efficient compared to training personal PEFT for each target user from scratch. For storage, instead of storing the entire set of matrices in LoRA for each target user, \ourmethod{} only needs to store the selected PEFT piece index $\mathcal{S}^l$ and corresponding weights $w_{s}^l$ across all layer positions,  ensuring \ourmethod{} storage efficient.

\section{Experiment Settings}

\begin{table*}[t]
  \centering
    \caption{Main experiment results on the LaMP benchmark. R-1 and R-L denote ROUGE-1 and ROUGE-L. $k$ refers to the number of retrieved items, with $k$=0 indicating no retrieval; $k$=1 is the default. $\uparrow$ means higher values are better, and $\downarrow$ means lower values are better. The best score for each task is in \textbf{bold}, and the second best is \underline{underlined}. `$^*$' indicates significant improvement against counterparts without \ourmethod{}.}
  \begin{adjustbox}{max width=1\linewidth}
    \begin{tabular}{llcccccccccc|ccc}
    \toprule[1.5pt]
    \multirow{2}{*}{\textbf{Task}} & \multirow{2}{*}{\textbf{Metric}} & \multicolumn{2}{c}{\textbf{Non-Personalized}} & \textbf{RAG} & \textbf{PAG} & \multicolumn{3}{c}{\textbf{PEFT Retrieval}} &\multicolumn{3}{c}{\textbf{\ourmethod{} (Ours)}} & \multicolumn{3}{c}{\textbf{OPPU}} \\
    \cmidrule(r){3-4} \cmidrule(r){5-5} \cmidrule(r){6-6} \cmidrule(r){7-9} \cmidrule(r){10-12}  \cmidrule(r){13-15}
    & & k=0 & Random & k=1 & k=1 & Base & +RAG & +PAG & Base & +RAG & +PAG & Base & +RAG & +PAG \\
    \midrule[0.75pt]

    \multirow{2}{*}{\makecell[l]{\textsc{LaMP-1: Personalized}\\\textsc{Citation Identification}}} & Acc $\uparrow$ & .536 & .576 & .584 & .656 & .480 &  .576 & .656 & .592$^*$ & .579 & \textbf{.672}$^*$ & .560 & .584 & \underline{.664} \\
    & F1 $\uparrow$&  .532 & .568 & .567 & .654 & .361 & .556 & .653 & .589$^*$ & .564 & \textbf{.664}$^*$ & .553 & .567 & \underline{.658} \\
    \hline
    
    \multirow{2}{*}{\makecell[l]{\textsc{LaMP-2: Personalized}\\\textsc{Movie Tagging}}} & Acc $\uparrow$& .340 & .301 & .417 & .469 & .336 & .419 & .477 & .410$^*$ & .452$^*$ & \underline{.499}$^*$ & .463 & .467 & \textbf{.507} \\
    & F1 $\uparrow$ & .268 & .255 & .331 & .375 & .265& .326& .380& .301$^*$ & .343$^*$ & \underline{.383}$^*$ & .320 & .349 & \textbf{.385} \\
    \hline
    \multirow{2}{*}{\makecell[l]{\textsc{LaMP-3: Personalized}\\\textsc{Product Rating}}} & MAE $\downarrow$& .645 & .336 & .301 & .301 & .431 & .305 & .299 & \textbf{.262}$^*$ & .272$^*$ & \textbf{.262}$^*$ & \textbf{.262} & .272 & \underline{.266} \\
    & RMSE $\downarrow$& 1.277 & .662 & .639 & .618 & .897 & .622 & .608 & \textbf{.558}$^*$ & .580$^*$ & \underline{.561}$^*$ & \textbf{.558} & .580 & \underline{.561}\\
    \hline
    \multirow{2}{*}{\makecell[l]{\textsc{LaMP-4: Personalized}\\\textsc{News Headline Gen.}}} & R-1 $\uparrow$& .175 & .179 & .201 & .204 & .189& .193 & .197 & .193$^*$ & \underline{.205}$^*$ & \underline{.205} & .193 & \underline{.205} & \textbf{.209}\\
    & R-L $\uparrow$& .158 & .162 & .183 & .184 & .171 & .175 & .178 & .174$^*$ & \underline{.186}$^*$ & \underline{.186} & .173 & .185 & \textbf{.190} \\
    \hline
    \multirow{2}{*}{\makecell[l]{\textsc{LaMP-5: Personalized}\\\textsc{Scholarly Title Gen.}}} & R-1 $\uparrow$& .485 & .486 & .501 & .505 & .488 & .508 & .509 & .488 & .510$^*$ & \textbf{.515}$^*$ & .490 & .509 & \underline{.512} \\
    & R-L $\uparrow$& .436 & .439 & .450 & .453 & .432 & .448 & .456 & .439 & .458$^*$ & \textbf{.460}$^*$ & .439 & .457 & \underline{.459} \\
    \hline    
    \multirow{2}{*}{\makecell[l]{\textsc{LaMP-7: Personalized}\\\textsc{Tweet Paraphrasing}}} & R-1 $\uparrow$ & .516 & .514  & .552 & \textbf{.565} & .522 & .552& .559 & .528$^*$ & \underline{.563}$^*$ & \textbf{.565} & .529 & .559 & .561 \\
    & R-L $\uparrow$& .463 & .457 & .511 & .517 & .475 & .512 & .517 & .482$^*$ & \textbf{.521}$^*$ & \underline{.519} & .480 & .515 & \underline{.519}\\
    \bottomrule[1.5pt]
    \end{tabular}%
    \end{adjustbox}
    \label{main_results}
\end{table*}


\paragraph{Datasets}
We adopt the Large Language Model Personalization (LaMP) benchmark \cite{salemi2023lamp} for our experiments, which consists of six public language model personalization tasks, including three text classification tasks (personalized citation identification, movie tagging, and producing rating) and three text generation tasks (personalized news headline generation, scholarly title generation, and tweet paraphrasing).\footnote{Task details can be found in Appendix \ref{task_details}. We exclude the LaMP-6: Email subject generation task since it involves private data that we cannot access.} We randomly select 25\% of users to train the base model for task adaptation. From the remaining users, we randomly sample 100 to serve as test users for efficient and fair comparison with OPPU \cite{tan2024democratizing}. The rest of the users are used as sharer candidates who consent to share their PEFT parameters.\footnote{Statistics are presented in Table \ref{tab:benchmark_stat}.}


\paragraph{Baselines}
We compare our proposed \ourmethod{} with the non-personalized baseline, prompt-based methods (retrieval-augmented \cite{salemi2023lamp} and profile-augmented personalization \cite{richardson2023integrating}), and PEFT-based personalization methods (PEFT retrieval \cite{zhao2024loraretriever} and OPPU \cite{tan2024democratizing}). Although PEFT retrieval has not been applied to personalization before, we employ it as a PEFT-level composition baseline. compared with \ourmethod{}, OPPU requires significantly more resources, which can be seen as the upper bound for sharer personal PEFT composition. We provide more baseline details in Appendix \ref{sec:baseline}. 
For all baselines and \ourmethod{}, we use \texttt{Llama-2-7B} \cite{touvron2023llama} as the base LLM and BM25 \cite{trotman2014improvements} for retrieval operations to ensure efficient and fair comparisons.

\paragraph{Evaluation Metrics}
Following LaMP \cite{salemi2023lamp}, we use accuracy and F1-score for personalized text classification tasks (LaMP-1 and LaMP-2), and MAE and RMSE for LaMP-3: personalized product rating. For personalized text generation tasks (LaMP-4, LaMP-5, and LaMP-7), we adopt ROUGE-1 and ROUGE-L \cite{lin-2004-rouge}. Higher scores indicate better performance for all metrics except RMSE and MAE used in LaMP-3.

\section{Results}
Table \ref{main_results} shows the performance on the curated test set of six public tasks in the LaMP benchmark. We have observations as follows.
\paragraph{Performance with \ourmethod{}.} Models equipped with \ourmethod{} outperform non-personalized, RAG, and PAG counterparts across all six tasks. In personalized text classification, \ourmethod{} achieves 11.79\% and 6.02\% relative gains in accuracy and F1-score for movie tagging, and 27.32\% and 24.92\% improvements in MAE and RMSE for product ratings. For personalized text generation, \ourmethod{} shows 4.25\% and 4.28\% relative improvements in ROUGE-1 and ROUGE-L scores for news headline generation. These results demonstrate \ourmethod{}'s effectiveness in enhancing LLM personalization.


\paragraph{\ourmethod{} \emph{vs.} PEFT Retrieval.} Compared to the PEFT retrieval method, \ourmethod{} shows clear superiority. For instance, \ourmethod{} achieves 8.76\% and 22.09\% performance gains in accuracy and F1-score for citation identification. Significant improvements are also seen in movie tagging, product rating prediction, and news headline generation tasks, highlighting the benefits of fine-grained PEFT piece composition over PEFT-level composition, which may risk user data leakage.
 
\paragraph{\ourmethod{} \emph{vs.} OPPU.} Compared to OPPU, which trains personal PEFT from scratch and requires more computational and storage resources, \ourmethod{} achieves comparable or slightly better results. Specifically, \ourmethod{} achieves 99.28\% of OPPU's performance on average with 7 times less computation and 38 times less storage in personalized text classification. In personalized text generation, \ourmethod{} shows comparable or better results in scholarly title generation and tweet paraphrasing.  

\paragraph{\ourmethod{} with Non-Parametric Knowledge.} Integrating both parametric user knowledge in personal PEFT and non-parametric in retrieval and user profile leads to notable performance gain. Averaging all tasks, RAG and PAG bring 1.67\% and 12.4\% performance gain in text classification tasks, as well as 6.11\% and 6.35\% enhancement in text generation tasks. 

Note that introducing RAG and PAG means users would expose their historical data or profiles to a centralized LLM, raising concerns about how user data are stored, used, and protected, and potentially affecting model ownership. For users prioritizing privacy and ownership, pure \ourmethod{} without retrieval avoids revealing user data to centralized LLM, and our experiments show it significantly outperforms non-personalized baselines. Conversely, those seeking optimal performance and consent to reveal data to centralized LLMs should opt for \ourmethod{}+RAG/PAG.

\begin{figure}[t]
    \centering
    \includegraphics[width=1\linewidth]{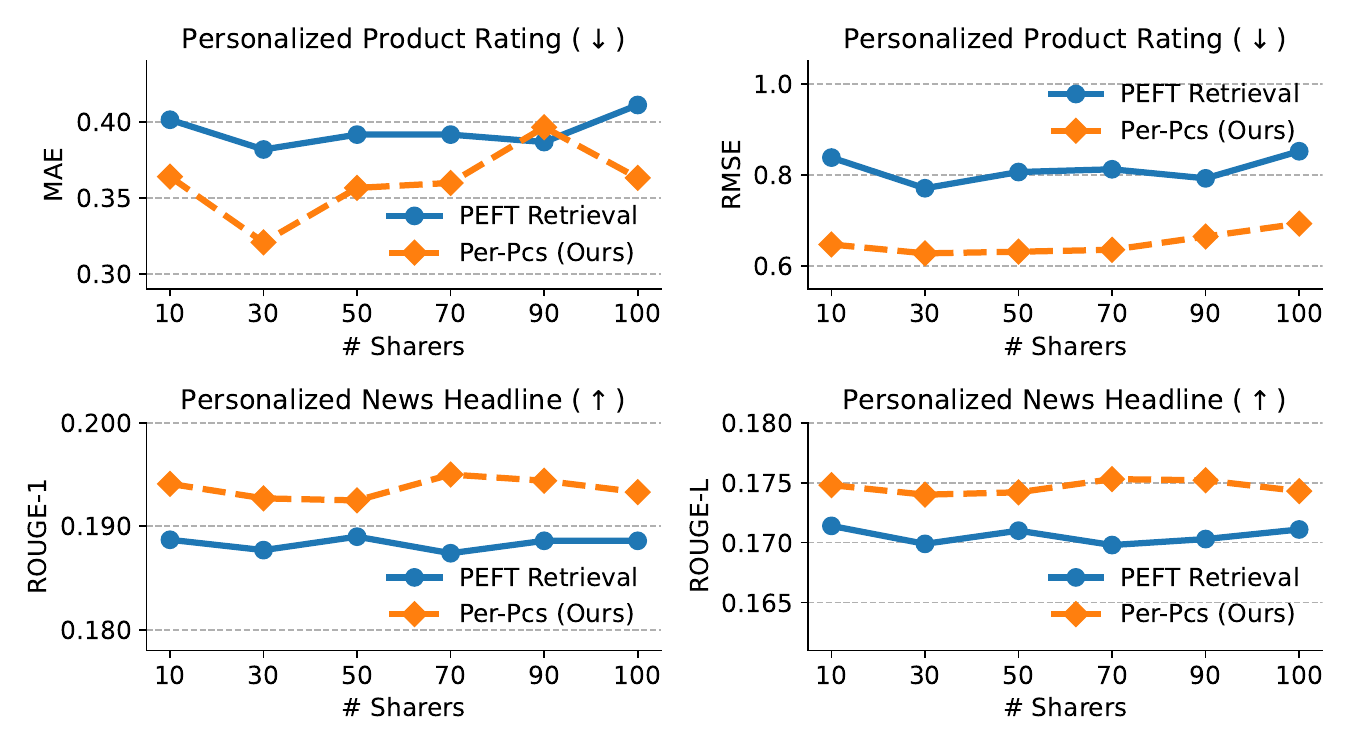}
    \caption{Model performance with different numbers of sharers in the product rating task (lower values indicate better performance). Our piece-level composition \ourmethod{} is stable and consistently outperforms the PEFT-level composition baseline.}
    \label{fig:anchor_number}
\end{figure}

\section{Analysis}
\paragraph{Robustness against Sharer Count}
In real-world deployment, the number of users who consent to share their personal PEFT can vary, and computational resources may constrain the number of sharers, making the sharer count a crucial factor in \ourmethod{}. In this experiment, we alter the number of sharers in two representative tasks from the text classification and generation categories to test the model's robustness. As shown in Figure \ref{fig:anchor_number}, \ourmethod{} exhibits relatively stable performance despite changes in the number of sharers and achieves the best performance with just 30 sharers in the personalized product rating prediction task, demonstrating its strong efficiency. Compared with the PEFT-level composition baseline (PEFT Retrieval), \ourmethod{} consistently shows better performance, highlighting the effectiveness of fine-grained piece-level composition in user modeling.

\begin{figure}[t]
    \centering
    \includegraphics[width=1\linewidth]{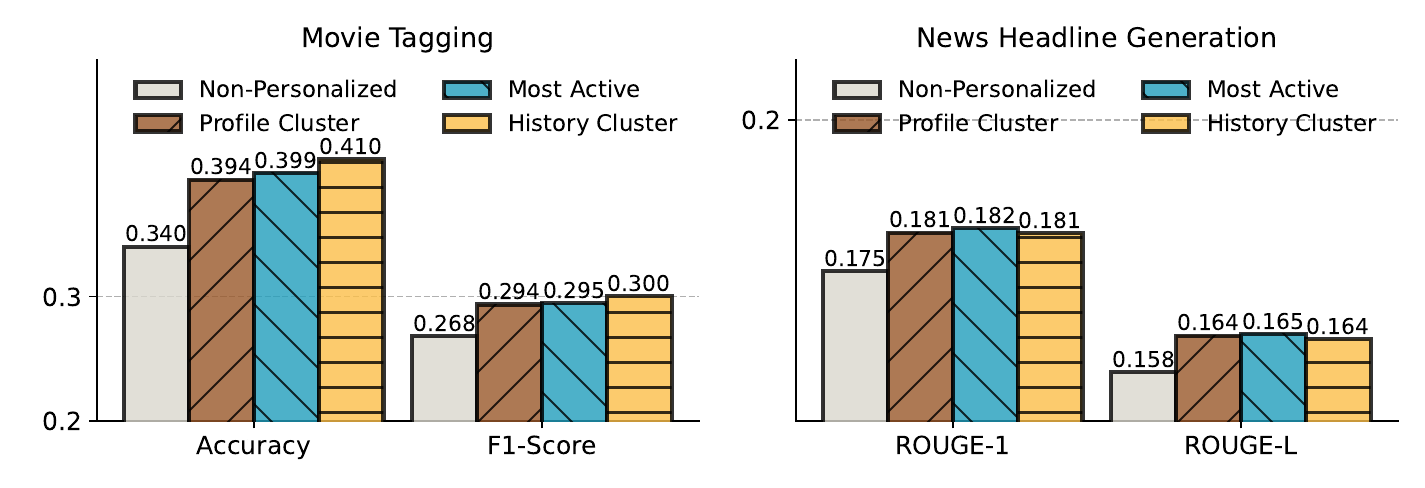}
    \caption{Performance of \ourmethod{} on movie tagging and news headline generation tasks with different sharer selection strategies. We find \ourmethod{} is robust to the choice of sharers.}
    \label{fig:anchor_selection}
\end{figure}

\begin{figure}[t]
    \centering
    \includegraphics[width=1\linewidth]{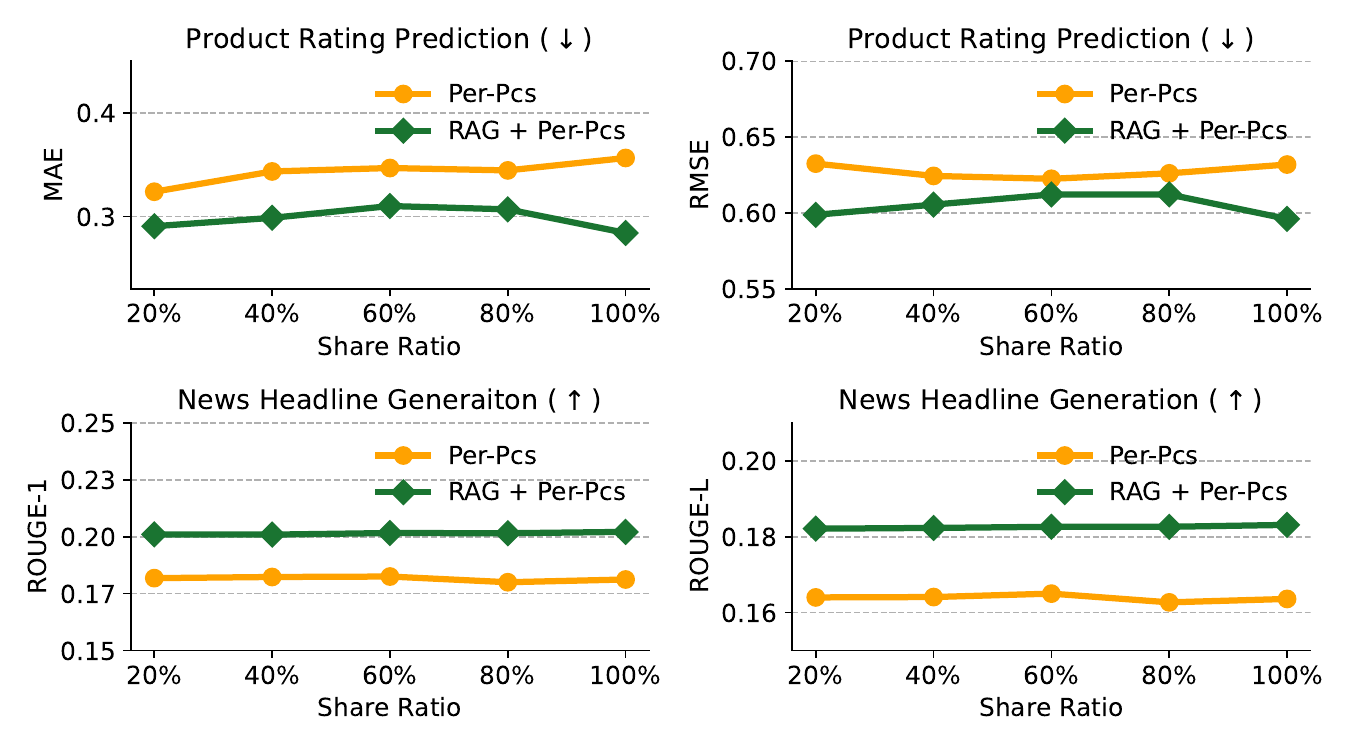}
    \caption{The \ourmethod{} performance with different PEFT parameter sharing ratios. \ourmethod{} maintains stable performance with a small sharing ratio, while non-parametric user knowledge via RAG enhances stability and performance.}
    \label{fig:share_ratio}
\end{figure}

\begin{figure*}[t]
    \centering
    \includegraphics[width=0.9\linewidth]{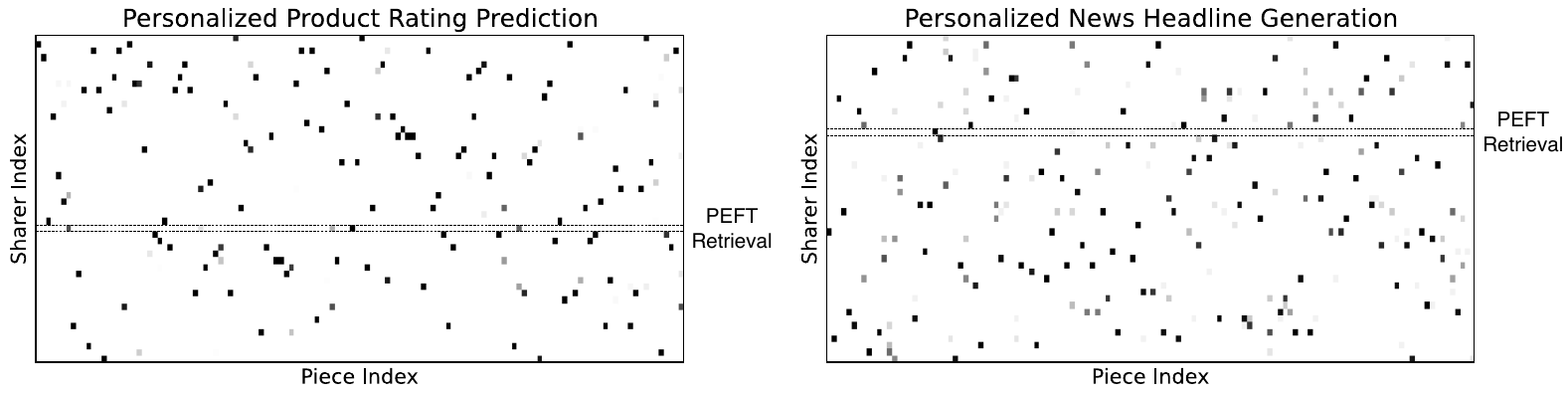}
    \caption{Case study on a specific user's PEFT assembled from sharers and corresponding piece weights in \ourmethod{}, compared with the PEFT retrieval choice. Unlike PEFT-level retrieval, \ourmethod{} models user history data in a more fine-grained manner while ensuring the privacy of sharers.}
    \label{fig:case_study}
\end{figure*}

\paragraph{On Sharer Selection Strategy}
\label{sec:anchor_selection}
In the main results, we present findings based on selecting sharers by clustering user history embeddings. However, users can have diverse distributions, and those who consent to share PEFT parameters may be biased in their distribution. Therefore, we tested \ourmethod{} with different sharer selection strategies to demonstrate its robustness against sharer selection. Specifically, we tested three strategies by restricting the sharer number to 50 users: "Most Active," which selects the 50 most active users; "Profile Cluster," which uses a \texttt{DeBERTa-v3-Large} encoder to obtain user embeddings for k-means clustering; and "History Cluster," the default setting, which averages user history embeddings to obtain user embeddings for clustering. As shown in Figure \ref{fig:anchor_selection}, all sharer selection strategies lead to better performance than the non-personalized baseline. Furthermore, \ourmethod{}'s performance remains stable across different sharer selection strategies, demonstrating its robustness. We hypothesize that fine-grained piece-level parameter composition can decompose complex user preferences from diverse dimensions, facilitating robustness against different sharer distributions.

\paragraph{Shared Pieces Ratio Study}
We designed \ourmethod{} to enable sharers to share a small portion of their PEFT parameters, preserving user privacy while maintaining strong performance. In this experiment, we varied the PEFT parameter sharing ratio and assessed its impact on model performance. Shown in Figure \ref{fig:share_ratio}, using two representative tasks from text classification and generation, we found that \ourmethod{} is highly robust to the sharing ratio, achieving comparable performance with just 20\% of the sharers' PEFT parameters compared to full parameter sharing. Additionally, with non-parametric user knowledge from RAG, \ourmethod{} demonstrates greater stability and performance. These results show that \ourmethod{} effectively balances privacy preservation and model performance.

\paragraph{Case Study}
To better understand the mechanism of piece-level composition in \ourmethod{}, we conducted a case study on piece selection and corresponding composition weights in product rating prediction and news headline generation, representing text classification and generation categories, respectively. As illustrated in Figure \ref{fig:case_study}, we observe that in both text classification and generation tasks, the selected pieces are diverse. Additionally, the weight distribution in generation tasks is more uniform, likely due to the intrinsic complexity of personality in text generation tasks.
Compared with PEFT-level retrieval, we find that \ourmethod{} almost never selects the same PEFT chosen by retrieval, yet it outperforms PEFT retrieval by 19.11\% and 4.15\% in product rating prediction and news headline generation tasks, respectively. We speculate that this is due to \ourmethod{}'s ability to effectively decompose and combine sharer PEFT pieces in a fine-grained manner, leveraging multiple sharers' parameters to enhance generalization.

\begin{figure}[t]
    \centering
    \includegraphics[width=1\linewidth]{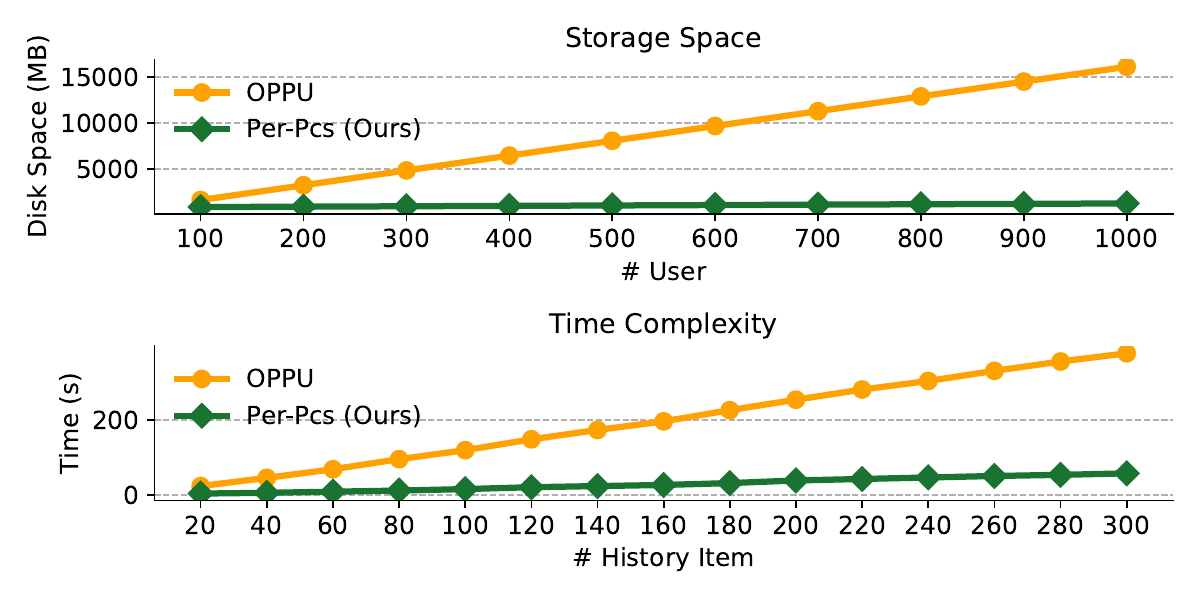}
    \caption{Comparison of storage and time complexity between our \ourmethod{} and OPPU, demonstrating that \ourmethod{} requires significantly less time to assemble personal PEFTs and less storage space to save them.}
    \label{fig:efficiency}
\end{figure}

\paragraph{Time and Space Complexity Analysis}

Scalability and efficiency are crucial for large-scale deployment of personalization methods. We compared \ourmethod{} and OPPU in terms of storage and assembly time. For storage efficiency, we used the product rating prediction task, observing requirements as the user count increased. For time efficiency, we examined a single user in the movie tagging task by varying the number of user history items. 
As shown in Figure \ref{fig:efficiency}, \ourmethod{} is significantly more efficient than OPPU in both storage and time. With increasing numbers of users and history items, \ourmethod{}'s efficiency advantage becomes even more pronounced, being approximately 38 times more efficient in storage and 7 times more efficient in time.
Moreover, as the number of users and history items grows, the efficiency advantage of \ourmethod{} becomes even more pronounced, being approximately 38 times more efficient in storage and 7 times more efficient in time.


\begin{figure}[t]
    \centering
    \includegraphics[width=1\linewidth]{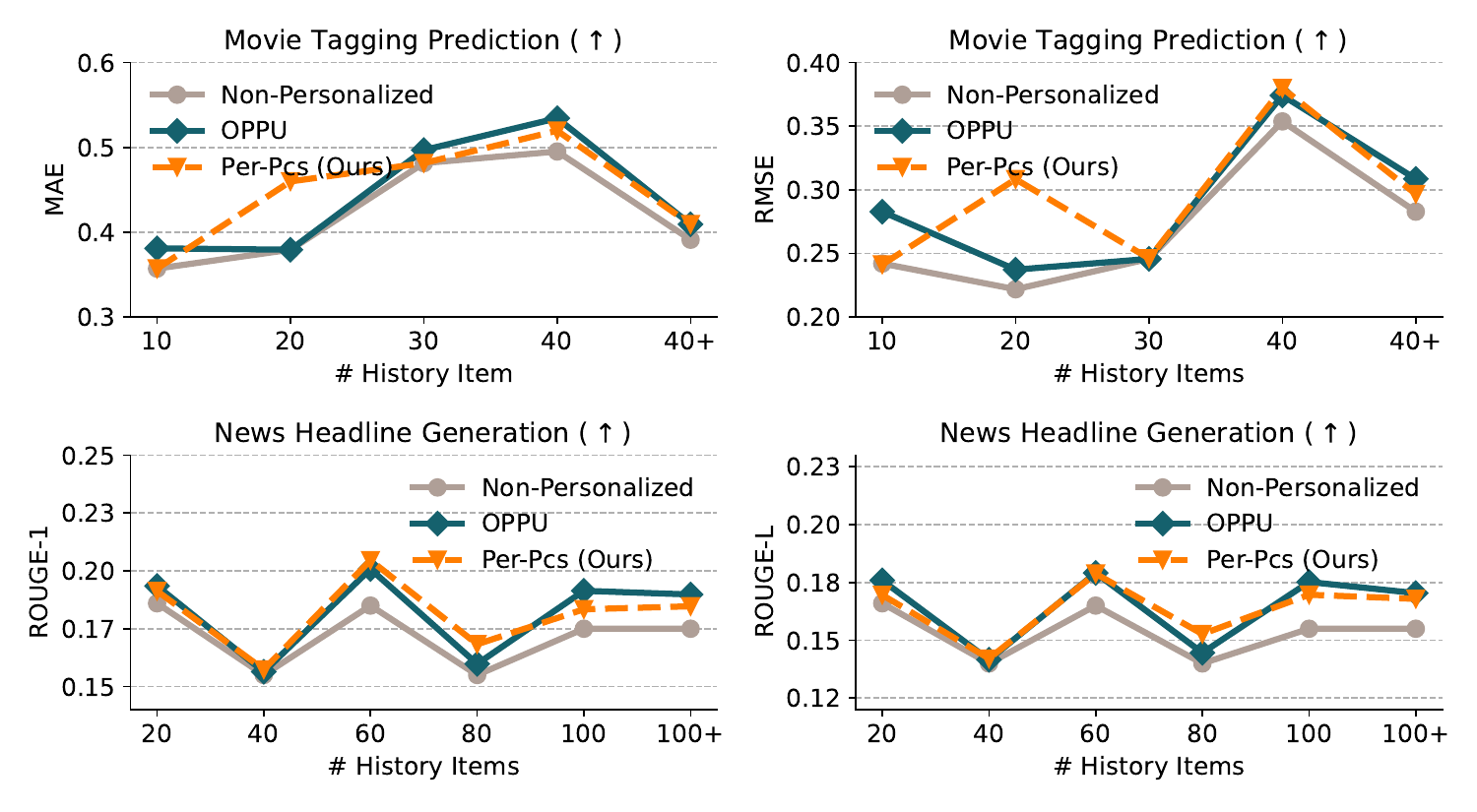}
    \caption{Model performance on personalized movie tagging and news headline generation for users with different numbers of history items.}
    \label{fig:active_level}
\end{figure}

\section{Modeling Users with Different Active Levels}
Users can exhibit different levels of activity, resulting in varying lengths of user history items for user modeling and personalization. To investigate the impact of user activity levels, quantified by the number of historical behavior items, on model performance, we randomly sampled 10 users from each range of activity levels. As shown in Figure \ref{fig:active_level}, we observe that (\textit{i}) \ourmethod{} generally shows stronger relative performance when user behavior items are fewer than 20, likely due to the collaborative signals captured during the assembling process that help the model understand user preferences. (\textit{ii}) \ourmethod{} generally performs similarly to OPPU, which requires training and maintaining personal PEFT from scratch and significantly more resources, and (\textit{iii}) both OPPU and \ourmethod{} outperform the non-personalized baseline at almost all activity levels. Overall, these results demonstrate the strong performance and robustness of \ourmethod{} across all user activity levels.

\section{Ablation Study}
As \ourmethod{} outperforms various baselines in personalization tasks, we investigate the impact of each design choice in \ourmethod{} to verify their effectiveness. More specifically, we perform ablation on the assembling process in both attention aggregation and piece selection steps. As is shown in Table \ref{tab:ablation}, the full \ourmethod{} outperforms all ablated models, proving our design choice's effectiveness. Moreover, the weighted aggregation of PEFT pieces has a significant impact on performance and is essential for model generalization for target users. We also find that TopP and TopK sampling strategies for pieces strategy would involve randomness and noises and eventually hurt the model performance. 

\begin{table}[t]
    \centering
    \caption{Performance of \ourmethod{} across different ablated versions: Top-p refers to setting a cumulative probability threshold $p$ and aggregating all pieces that first reach this threshold. Topk-Sampling denotes sampling one piece from the top $k$ pieces with normalized scores as probabilities.}
    \resizebox{1\linewidth}{!}{
    \begin{tabular}{l c c c c}
        \toprule[1.5pt] 
         \multirow{2}{*}{\textbf{Ablation Settings}} & \multicolumn{2}{c}{\textbf{LaMP-2}} & \multicolumn{2}{c}{\textbf{LaMP-4}} \\
         \cmidrule(r){2-3} \cmidrule(r){4-5} 
         & Acc & F1 & R-1 & R-L \\
        \midrule[0.75pt]
        
       full model & 0.410 & 0.301& 0.193& 0.174\\
       \hline
       w/o attention & 0.340 & 0.266 & 0.176 & 0.158\\
       replace Topk-Agg. w/ Topp-Agg. & 0.390 & 0.288 & 0.169 & 0.153\\
       replace Topk-Agg. w/ Topk-Sampling & 0.383 & 0.297 & 0.172 & 0.155 \\
       
       \bottomrule[1.5pt] 
    \end{tabular}
    }
    \label{tab:ablation}
\end{table}

\section{Related Work}
\subsection{Personalization of LLMs}
Existing LLM personalization methods can be categorized into prompt-based and Parameter Efficient Fine-tuning (PEFT)-based methods. 

\textit{Prompt-based personalization} method focuses on designing prompts that incorporate user-generated content and behavior to help LLMs understand user preferences, which can be further categorized into vanilla personalized prompting, retrieval-augmented personalized prompting, and profile-augmented personalized prompting.
Vanilla personalized prompting leverages LLMs' in-context learning and few-shot learning abilities by encoding either complete or randomly sampled user history behaviors as contextual examples \cite{dai2023uncovering, wang2023learning, kang2023llms}. To manage the rapidly growing user behavior and LLMs' limited context window, researchers have proposed retrieval-augmented methods for personalized LLMs \cite{salemi2023lamp}, and enhance the calibration \cite{mysore2023pearl} and optimize retrieval \cite{salemi2024optimization}. Moving beyond simple retrieval, some researchers have proposed profile-augmented personalization prompting, summarizing natural language user preferences and behavior patterns to augment user queries \cite{richardson2023integrating}, and constructing hierarchy personalized retrieval databases \cite{sun2024persona}. 

\textit{PEFT-based personalization} methods store user preferences and behavior patterns in parameters. OPPU \cite{tan2024democratizing} equips each user with a personal PEFT module, storing preferences in PEFT parameters and offering better generalization of user behavior patterns compared to prompt-based methods.
Another line of work focuses on designing personalized alignment methods via parameter merging \cite{jang2023personalized}, personalized RLHF \cite{li2024personalized, park2024principled}, personalized reward models \cite{cheng2023deserves}, and black-box LLM personalization \cite{zhuang2024hydra}.

\subsection{Model Parameter Composition}
Existing work has shown that performing weighted linear interpolation of model parameters leads to the composition of each model ability \cite{li2022branch, tam2023merging, dou2024avoiding}. This approach recycles efforts and computational resources used to create specialized models. These methods can be divided into model-, PEFT-, and piece-level compositions.

\textit{Model-level composition} methods treat the entire model parameter as the minimum composition unit \cite{wortsman2022model, zhang-etal-2024-working-memory, choshen2022fusing, rame2023model, jin2022dataless, liu2024towards, jian-etal-2024-expedited}. 
\citet{ilharco2022editing} propose the task vector, which subtracts the weights of a fine-tuned model from the pre-trained weights and conducts task vector arithmetic to enable generalization across tasks and domains. 
PEFT offers lightweight alternatives for fine-tuning LLMs by updating small, plug-in parameters while keeping the pre-trained weights frozen to save computational resources \cite{he2021towards}. 
In \textit{PEFT-level composition}, the entire PEFT module is treated as the minimum unit. By composing PEFT parameters, models can achieve task and domain generalization \cite{shah2023ziplora, gou2023mixture, zhang2023composing}. LoRAHub \cite{huang2023lorahub} uses a black-box optimizer to integrate specialized LoRAs, facilitating generalization to unseen tasks. Another line of work focuses on retrieving PEFT \cite{jang2023exploring}. LoRARetriever \cite{zhao2024loraretriever} retrieves and composes multiple LoRAs based on the given input.
In \textit{piece-level composition}, the minimum composition unit is a plug-in sub-component of PEFT within a specific layer. For instance, in LoRA, each low-rank update at a linear layer constitutes a "piece." \citet{muqeeth2024learning} focuses on task generalization and proposes recycling PEFT pieces by employing per-token and per-piece composition under zero-shot settings.


In this work, we propose \ourmethod{}, a personal PEFT sharing framework that takes advantage of piece-wise parameter composition, enabling users to share partial parameters. This approach ensures the sharer's privacy while maintaining model ownership, fine-grained user modeling, and strong efficiency for personalized LLM democratization. 

\section{Conclusion}

We proposed \ourmethod{}, a novel framework that enables users to share their personal PEFTs, creating community value while preserving privacy. For target users, \ourmethod{} maintained model ownership, efficiency, and fine-grained personalization by employing piece-level composition based on user history data. Extensive experiments showed that \ourmethod{} outperforms non-personalized and PEFT retrieval methods, and performs close to OPPU with significantly lower computational and storage resources.  We envisioned \ourmethod{} as a community-driven effort to advance personalized LLM, making it more modular, effective, and widely accessible.


\section{Limitations}
We identify two key limitations in \ourmethod{}. First, constrained by the dataset, our focus is primarily on one specific task per user rather than examining user behaviors across multiple tasks and domains. For instance, in the movie tagging task, users are solely engaged in that specific activity, without the inclusion of behaviors from other domains or platforms. Despite this, the \ourmethod{} framework is inherently adaptable to any text sequence generation task and is compatible with diverse user instructions across various tasks and domains. Personalizing LLM across a broader range of tasks and domains is left as future work.
Second, despite our proposed \ourmethod{} is compatible with all PEFT methods that introduce trainable modules throughout the model, such as Adapter \cite{houlsby2019parameter}, $\mathrm{(IA)^3}$ \cite{liu2022few}, and prefix tuning \cite{li-liang-2021-prefix}, we primarily focus on LoRA in this work. This is due to LoRA's popularity, widespread use, and superior performance demonstrated by OPPU \cite{tan2024democratizing}, while we expect to expand our experiment and analysis to more PEFT methods in future work.


\section{Ethical Considerations}
\paragraph{Data Bias} Personalizing LLMs relies heavily on personal data input into the system. If this data is biased or unrepresentative, the model's outputs could perpetuate these biases, leading to unfair or prejudiced responses. It is crucial to monitor and mitigate such biases in personal data and personalized models to ensure fair, unbiased, and safe responses from personalized LLMs. In \ourmethod{}, where users build personal PEFTs through collaborative efforts, bias in user data could spread within the community, amplifying negative effects. Future work could focus on preventing harmful biases in user data at both the personal and community levels.

\paragraph{Accessibility} While advancing personalized LLMs aims to enhance user interactions with AI systems, their complexity and resource-intensive nature can pose accessibility challenges. Smaller entities or individual researchers with limited computational power and budgetary constraints may struggle to engage with advanced personalized LLMs, potentially widening the gap in AI research and application. Efforts should be made to make these technologies more accessible to a broader audience to ensure equitable advancement in AI research.

\paragraph{Privacy}
The personalization of LLMs necessitates tailoring responses based on user-specific data, which may include sensitive or private information. The ability of an LLM to adapt its outputs to individual users raises privacy concerns, as it might inadvertently reveal personal details. This highlights the importance of implementing robust privacy safeguards in LLM personalization to ensure that personal data is handled respectfully and securely, preventing any unintended disclosures.

\section{Acknowledgements}

This work was supported by NSF IIS-2119531, IIS-2137396, IIS-2142827, IIS-2234058, CCF-1901059, and ONR N00014-22-1-2507.


\bibliography{custom}
\clearpage

\clearpage
\appendix

\section{Computation Resources Details}
All experiments are implemented on a server with 3 NVIDIA A6000 GPU and Intel(R) Xeon(R) Silver 4210R CPU @ 2.40GHz with 20 CPU cores.

\begin{table*}[t]
    \centering
    \caption{Hyperparameter settings of \ourmethod{} across six tasks on LaMP data.}
    \resizebox{0.7\linewidth}{!}{
    \begin{tabular}{l c c c c c c c c}
        \toprule[1.5pt] 
       \multirow{2}{*}{\textbf{Task}}  & \multicolumn{3}{c}{\textbf{Sharer PEFT}} & \multicolumn{3}{c}{\textbf{Sharer Gate}} & \multicolumn{2}{c}{\textbf{\ourmethod{} Assemble}}\\
       \cmidrule(r){2-4} \cmidrule(r){5-7} \cmidrule(r){8-9}
         & batch size  & epoch & lr & batch size & step & lr &  top-k & batch size \\
        \midrule[0.75pt] 
      \makecell[l]{\textsc{LaMP-1: Personalized}\\\textsc{Citation Identification}} & 16 & 1 & 1e-5 & 6 & 100 & 1e-5 & 1 & 16\\
      \hline
      \makecell[l]{\textsc{LaMP-2: Personalized}\\\textsc{Movie Tagging}} & 6 & 3& 2e-5 & 6 & 100 & 2e-5 &  3 & 16 \\
      \hline
      \makecell[l]{\textsc{LaMP-3: Personalized}\\\textsc{Product Rating}} & 2 & 2 & 1e-5 & 4 & 100 & 1e-5 & 1 & 6\\
      \hline
      \makecell[l]{\textsc{LaMP-4: Personalized}\\\textsc{News Headline Gen.}} & 10 & 3 & 2e-5 & 6 & 50 & 2e-5 & 1 & 16\\
      \hline
      \makecell[l]{\textsc{LaMP-5: Personalized}\\\textsc{Scholarly Title Gen.}} & 3 & 2 & 2e-5 & 6 & 50 & 2e-5 & 1 & 10\\
      \hline
      \makecell[l]{\textsc{LaMP-7: Personalized}\\\textsc{Tweet Paraphrasing}} & 16 & 2 & 1e-5 & 6 & 50 & 2e-5 & 2 & 16 \\
       \bottomrule[1.5pt] 
    \end{tabular}
    }
    
    \label{tab:hyper}
\end{table*}

\section{Hyperparameters}
The hyperparameters of \ourmethod{} are presented in Table \ref{tab:hyper} to facilitate further research.

\section{Scientific Artifacts}
\ourmethod{} is built with the help of many existing scientific artifacts, including PyTorch \cite{paszke2019pytorch}, Numpy \cite{harris2020array}, huggingface, and transformers \cite{wolf2020transformers}. We will make the \ourmethod{} implementation publicly available to facilitate further research.

\section{Task Details}
\label{task_details}
We present the task details as follows to help readers gain a better understanding of the task format.
\begin{itemize}[leftmargin=*]
    \item \textbf{Personalized Citation Identification} is a binary text classification task. Specifically, given user $u$ writes a paper $x$, the task aims to make the model determine which of the two candidate papers $u$ will cite in paper $x$ based on the user's history data, which contains the publications of user $u$.

    \item \textbf{Personalized News Categorization} is a 15-way text classification task to classify news articles written by a user $u$. Formally, given a news article $x$ written by user $u$, the language model is required to predict its category from the set of categories based on the user's history data, which contains the user's past article and corresponding category.

    \item \textbf{Personalized Movie Tagging} is a 15-way text classification task to make tag assignments aligned with the user's history tagging preference. Specifically, given a movie description $x$, the model needs to predict one of the tags for the movie $x$ based on the user's historical movie-tag pairs.

    \item \textbf{Personalized Product Rating} is a 5-way text classification task and can also be understood as a regression task. Given the user $u$'s historical review and rating pairs and the input review $x$, the model needs to predict the rating corresponding to $x$ selected from 1 to 5 in integer. 

    \item \textbf{Personalized News Headline Generation} is a text generation task to test the model's ability to capture the stylistic patterns in personal data. Given a query $x$ that requests to generate a news headline for an article, as well as the user profile that contains the author's historical article-title pairs, the model is required to generate a news headline specifically for the given user.

    \item \textbf{Personalized Scholarly Title Generation} is a text generation task to test personalized text generation tasks in different domains. In this task, we require language models to generate titles for an input article $x$, given a user profile of historical article-title pairs for an author.

    \item \textbf{Personalized Tweet Paraphrasing} is also a text generation task that tests the model's capabilities in capturing the stylistic patterns of authors. Given a user input text $x$ and the user profile of historical tweets, the model is required to paraphrase $x$ into $y$ that follows the given user's tweet pattern.
\end{itemize}

\begin{table}[t]
    \centering
    \caption{Dataset statistics: We report average sequence length in terms of number of tokens. \#Q is the number of queries, L$_{in}$ and L$_{out}$ are the average length of input and output sequence respectively, and \#History is the number of user history items. To save space, task names can be found in Table~\ref{main_results}.}
    \begin{adjustbox}{max width=1\linewidth}
    \begin{tabular}{l r r r r r r r r}
         \toprule[1.5pt]
        \multirow{2}{*}{\makecell[l]{\textbf{Task in}\\ \textbf{LaMP}}} & \multicolumn{4}{c}{\textbf{Sharer Candidates}} & \multicolumn{4}{c}{\textbf{Target Users}}\\
         \cmidrule(r){2-5} \cmidrule(r){6-9}
         &\#Q & \#History & L$_{in}$ & L$_{out}$ &  \#Q & \#History & L$_{in}$ & L$_{out}$ \\
         \midrule[0.75pt]
         \textbf{1}&5,334 & 88.5 & 51.4 & 1.0 & 125 & 147.2 & 50.8 & 1.0 \\
         \textbf{2} & 2,385 & 12.3 & 92.5 & 1.7 & 2,228 & 37.3 & 92.3 & 2.0 \\
         \textbf{3}& 15,034 & 202.5&132.1 & 1.0 & 614&360.6 & 160.9 & 1.0 \\
         \textbf{4}& 7,568 & 31.3& 30.1 & 10.1 & 3,949 & 155.9& 26.5& 10.7\\
         \textbf{5}& 10,821 & 94.3 & 162.7 & 9.7 & 608 & 144.0 & 158.9 & 9.7 \\
         \textbf{7}& 9,978 & 15.7 & 299.6 & 16.9 & 114 & 77.2 & 30.3 & 17.0\\
         \bottomrule[1.5pt]
    \end{tabular}
    \end{adjustbox}
    \label{tab:benchmark_stat}
\end{table}

\section{Baseline Details}
\label{sec:baseline}
\begin{itemize}[leftmargin=*]
\item \textbf{Non-Personalized baseline}:
We present two approaches under the non-personalized setting: non-retrieval and random history. \textit{Non-retrieval method} ($k$=0) refers to only feeding the user's query without revealing the user's behavior history to the LLMs. \textit{Random history} baseline means augmenting the user's query with random history behavior from all user history corpus.

\item \textbf{Retreival-Augmented Personalization (RAG)}: We follow the retrieval-augmented personalization method presented in LaMP \cite{salemi2023lamp}, where the user's query is augmented with top $k$ retrieved items from the corresponding user's history corpus. We take $k$=1 by default in this work.

\item \textbf{Profile-Augmented Personalization (PAG)}: This method is taken from \citet{richardson2023integrating}, in which the user's input sequence would concatenate the user's profile summarizing the user's preference and behavior patterns. In our experiments, we generate user profiles using the \texttt{Mistral-7B} \cite{jiang2023mistral} model. Moreover, the profile-augmented method could be combined with the retrieval augmentation. In this case, we take the number of retrieval items $k$=1 following the setting of \citet{richardson2023integrating}.

\item \textbf{PEFT Retrieval}: Similar to \citet{jang2023exploring, zhao2024loraretriever}, when a target user comes, we compute the cosine similarity between embeddings of target users and sharers and find the top-k similar users and conduct weighted aggregation to obtain target user's PEFT. The PEFT retrieval method has not been applied to LLM personalization before and we select it as a PEFT-level composition baseline to compare with our proposed fine-grained piece-level composition method.

\item \textbf{OPPU}: This method was proposed by \citet{tan2024democratizing}, which trains a PEFT for each user from scratch and can be integrated with prompt-based personalization methods. Compared to our \ourmethod{}, OPPU requests significantly more computation and storage.
\end{itemize}

\section{Dataset Statistics}

The dataset statistics are presented in Table \ref{tab:benchmark_stat}.

\section{Prompt Details}
\label{sec:profile_gen}

We present the prompt used in our experiments in this section, where the text in \texttt{\{BRACES\}} can be replaced with content specific to different users and queries. Prompts for user profile generation are presented in Table \ref{tab:profile_prompt}, prompts for personalization tasks are presented in Table \ref{tab:task_prompt}

\begin{table*}[t]
\caption{Prompt for user profile generation.}
\begin{adjustbox}{max width=1\linewidth}
\begin{tabular}{p{3in}p{5in}}
    \toprule[1.5pt]
    {\textbf{Task}}   & {\textbf{Prompt}}  \\
     \midrule[1pt]
         
    \textsc{LaMP-1: Personalized Citation Identification}
    &
    Write a summary, in English, of the research interests and topics of a researcher who has published the following papers. Only generate the summary, no other text. User History: \texttt{\{USER HISTORY\}} Answer:\\
    \hline

    \textsc{LaMP-2: Personalized Movie Tagging}&
    Look at the following past movies this user has watched and determine the most popular tag they labeled. Answer in the following form: most popular tag: <tag>. User History: \texttt{\{USER HISTORY\}} Answer:\\
    \hline
    \textsc{LaMP-3: Personalized Product Rating} & Based on this user\'s past reviews, what are the most common scores they give for positive and negative reviews? Answer in the following form: most common positive score: <most common positive score>, most common negative score: <most common negative score>. User History: \texttt{\{USER HISTORY\}} Answer:\\
    \hline

    \textsc{LaMP-4: Personalized News Headline Generation} & Given this author\'s previous articles, try to describe a template for their headlines. I want to be able to accurately predict the headline gives one of their articles. Be specific about their style and wording, don\'t tell me anything generic. User History: \texttt{\{USER HISTORY\}} Answer:\\
    \hline
    
    \textsc{LaMP-5: Personalized Scholarly Title Generation} & Given this author\'s previous publications, try to describe a template for their titles. I want to be able to accurately predict the title of one of the papers from the abstract. Only generate the template description, nothing else. User History: \texttt{\{USER HISTORY\}} Answer:
    \\
    \hline

    \textsc{LaMP-7: Personalized Tweet Paraphrasing} & 
    Given this person\'s previous tweets, try to describe a template for their tweets. I want to take a generic sentence and rephrase it to sound like one of their tweets, with the same style/punctuation/capitalization/wording/tone/etc. as them. Only give me the template description, nothing else. User History: \texttt{\{USER HISTORY\}} Answer:\\
     
     \bottomrule[1.5pt]
\end{tabular}
\end{adjustbox}
\label{tab:profile_prompt}
\end{table*}

\begin{table*}[t]
\caption{Prompt for personalization tasks.}
\begin{adjustbox}{max width=1\linewidth}
\begin{tabular}{p{3in}p{5in}}
    \toprule[1.5pt]
    {\textbf{Task}}   & {\textbf{Prompt}}  \\
     \midrule[1pt]
         
    \textsc{LaMP-1: Personalized Citation Identification}
    &
    \noindent\#\#\# User Profile:

    \noindent\texttt{\{USER PROFILE\}}
    
    \#\#\# User History:
    
    \noindent\texttt{\{USER HISTORY\}}

    \noindent \#\#\# User Instruction:
    
    \noindent Identify the most relevant reference for the listed publication by the researcher. Select the reference paper that is most closely related to the researcher\'s work. Please respond with only the number that corresponds to the reference.
    
    \noindent Paper Title: \texttt{\{QUERY PAPER TITLE\}} Reference: [1] - \texttt{\{OPTION1\}} [2] - \texttt{\{OPTION2\}}
    
    \noindent Answer:\\
    \hline

    \textsc{LaMP-2: Personalized Movie Tagging}&
    \noindent\#\#\# User Profile:

    \noindent\texttt{\{USER PROFILE\}}
    
    \#\#\# User History:
    
    \noindent\texttt{\{USER HISTORY\}}

    \noindent \#\#\# User Instruction:
    
    Which tag does this movie relate to among the following tags? Just answer with the tag name without further explanation. tags: [sci-fi, based on a book, comedy, action, twist ending, dystopia, dark comedy, classic, psychology, fantasy, romance, thought-provoking, social commentary, violence, true story]
    
    Description: \texttt{\{QUERY MOVIE DESCRIPTION\}} Tag:\\
    \hline
    \textsc{LaMP-3: Personalized Product Rating} & \noindent\#\#\# User Profile:

    \noindent\texttt{\{USER PROFILE\}}
    
    \#\#\# User History:
    
    \noindent\texttt{\{USER HISTORY\}}

    \noindent \#\#\# User Instruction:

    What is the score of the following review on a scale of 1 to 5? just answer with 1, 2, 3, 4, or 5 without further explanation.
    
    Review: \texttt{\{QUERY REVIEW\}} Score:\\
    \hline

    \textsc{LaMP-4: Personalized News Headline Generation} & \noindent\#\#\# User Profile:

    \noindent\texttt{\{USER PROFILE\}}
    
    \#\#\# User History:
    
    \noindent\texttt{\{USER HISTORY\}}

    \noindent \#\#\# User Instruction:
    
    Generate a headline for the following article.
    
    Article: \texttt{\{QUERY ARTICLE\}} Headline:\\
    \hline
    
    \textsc{LaMP-5: Personalized Scholarly Title Generation} & \noindent\#\#\# User Profile:

    \noindent\texttt{\{USER PROFILE\}}
    
    \#\#\# User History:
    
    \noindent\texttt{\{USER HISTORY\}}

    \noindent \#\#\# User Instruction:
    
    Generate a title for the following abstract of a paper.
    
    Abstract: \texttt{\{QUERY ABSTRACT\}} Title:
    \\
    \hline

    \textsc{LaMP-7: Personalized Tweet Paraphrasing} & 
    \noindent\#\#\# User Profile:

    \noindent\texttt{\{USER PROFILE\}}
    
    \#\#\# User History:
    
    \noindent\texttt{\{USER HISTORY\}}
    
    \noindent \#\#\# User Instruction:
    
    Paraphrase the following text into tweet without any explanation before or after it.
    
    Text: \texttt{\{QUERY TEXT\}} Tweet:\\

     \bottomrule[1.5pt]
\end{tabular}
\end{adjustbox}
\label{tab:task_prompt}
\end{table*}

\end{document}